\documentclass[letterpaper, 10 pt, conference]{ieeeconf}  % Comment this line out if you need a4paper

\IEEEoverridecommandlockouts                              % This command is only needed if 
\usepackage{cite}
\usepackage{amsmath,amssymb,amsfonts}
\usepackage{graphicx}
\usepackage{textcomp}
\usepackage{xcolor}
\usepackage{hyperref}
\usepackage{url}
\title{\LARGE \bf
Long-Distance Gesture Recognition using Dynamic Neural Networks
}

\author{Shubhang Bhatnagar$^{1,2*}$, Sharath Gopal$^{2}$, Narendra Ahuja$^{1}$, Liu Ren$^{2}$% <-this % stops a space
\thanks{ $^{1}$ Department of Electrical and Computer Engineering, University of Illinois Urbana-Champaign.  % <-this % stops a space
  {\tt\small E-mails: \{sb56, n-ahuja\}@illinois.edu}} 
  \thanks{ $^{2}$ Bosch Research North America - Bosch Center for Artificial Intelligence. % <-this % stops a space
{\tt\small E-mails: \{Sharath.Gopal, Liu.Ren\}@us.bosch.com}
 }%
\thanks{ $^{*}$ Work done as an intern at Bosch Research.}
}
\begin{document}

%\thanks{*This work was not supported by any organization}% <-this % stops a space
%\thanks{ $^{1}$ Department of ECE, UIUC. Work done as an intern at Bosch Research }% <-this % stops a space

\maketitle
\thispagestyle{empty}
\pagestyle{empty}

%%%%%%%%%%%%%%%%%%%%%%%%%%%%%%%%%%%%%%%%%%%%%%%%%%%%%%%%%%%%%%%%%%%%%%%%%%%%%%%%

\begin{abstract}
Gestures form an important medium of communication between humans and machines. An overwhelming majority of existing gesture recognition methods are tailored to a scenario where humans and machines are located very close to each other. This short-distance assumption does not hold true for several types of interactions, for example gesture-based interactions with a floor cleaning robot or with a drone. Methods made for short-distance recognition are unable to perform well on long-distance recognition due to gestures occupying only a small portion of the input data. Their performance is especially worse in resource constrained settings where they are not able to effectively focus their limited compute on the gesturing subject. We propose a novel, accurate and efficient method for the recognition of gestures from longer distances. It uses a dynamic neural network to select features from gesture-containing spatial regions of the input sensor data for further processing. This helps the network focus on features important for gesture recognition while discarding background features early on, thus making it more compute efficient compared to other techniques. We demonstrate the performance of our method on the LD-ConGR long-distance dataset where it outperforms previous state-of-the-art methods on recognition accuracy and compute efficiency.
\end{abstract}

%\begin{IEEEkeywords}
%component, formatting, style, styling, insert
%\end{IEEEkeywords}

\section{Introduction}\label{sec:intro}
Gestures are an efficient non-verbal method for directing traffic from a distance, interacting in public places and communicating with the deaf. Gestures are an important part of human-machine interaction (HMI) where they provide a natural and friendly way to interact with robots and machines\cite{gesture_survey,robot_interact_gesture, guo2021human} at homes, offices, airports, hospitals and in automobiles. Recent advances in AR/VR and fitness technologies have increased the use of wearable devices (headsets, smart watches, health trackers) where hand, face and eye gestures\cite{gaze_gesture, vr_gesture, egogesture} are a contact-less way to collect user input and intent.

\begin{figure}[t!]
	 \centering
    \includegraphics[width=0.47\textwidth]{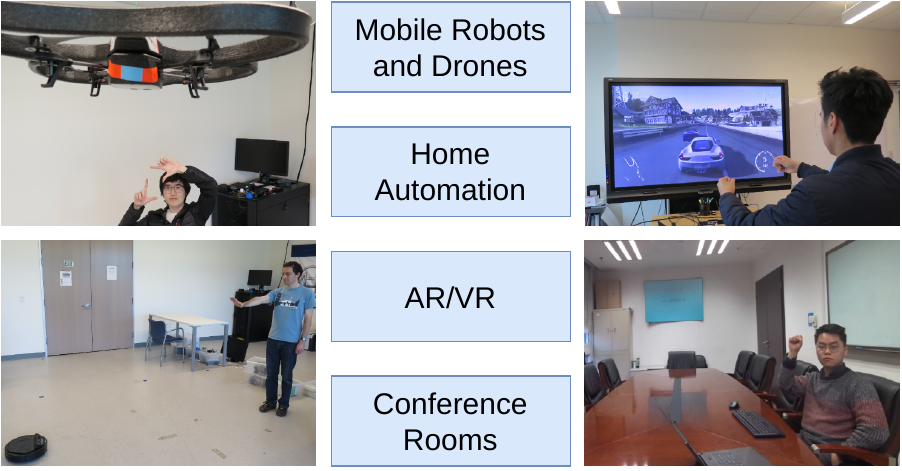}
    \caption{Example applications of our proposed method for long-distance gesture recognition.}
    \label{fig:long_distance_applications}
\end{figure}

Gesture recognition can be broadly categorized \cite{LDCon_GR} into short-distance and long-distance based on the distance between the sensors and the gesturing subject. Short-distance applications, such as car infotainment systems and desktops/laptops, have the subject within a 1m distance from the sensors. Long-distance applications include service robots (floor cleaning, warehouse, lawn mowing), conference/meeting rooms, home automation and IoT devices (adjust lights, TV controls),  AR/VR and video games, where the subject is typically far away (1-4m) from the sensors. In some of these applications, such as the mobile floor cleaning robot, short-distance recognition is not an option due to a very low camera position which forces the subject to stand farther away to be in its field-of-view (FOV). Fig \ref{fig:long_distance_applications} illustrates some of these long-distance applications. 

Self-occlusions, finger similarity and uncertainty of gesture duration are some of the common challenges faced by gesture recognition methods. Deployment of such methods in resource and compute constrained robots and devices requires us to consider the computational complexity and efficiency, in addition to accuracy, of the method. Long-distance scenarios pose an additional challenge of a large FOV in which the hands appear small, making it harder to recognize the gesture. Longer distances lead to reduction in signal strength and increased blur for the gesture, which further deteriorate performance. 

We address these challenges by proposing a new method that uses modern convolutional neural networks (CNNs) for recognizing hand gestures from visual data (RGB video). We leverage the efficiency and structural adaptability of dynamic neural networks \cite{dynamic_orig1, dynamic_orig2} for long-distance applications. 

The main contributions of this paper are:
\begin{itemize}
\item A novel dynamic neural network based gesture recognition method for long-distance applications.
\item Extensive experiments and evaluation on a long-distance dataset. We show that our method is 3.5\% more accurate and 28.5\% more compute-efficient compared to the previous best method (Table \ref{tab:results_main}).
\end{itemize}
The rest of the paper is organized as follows: We review previous work in gesture recognition and dynamic neural networks in Section \ref{sec:related_work}. Section \ref{sec:method} describes the details of our method and Section \ref{sec:exp} presents a thorough evaluation and analysis of our method compared to previous state-of-the-art techniques. Finally, we summarize and conclude in Section \ref{sec:conclusion}.

\section{Related Work}\label{sec:related_work}

\subsection{Gesture Recognition}
Gesture recognition methods are typically based on input from motion, visual or range sensors of human bodies, hands and legs. In this section, we focus on reviewing hand gesture recognition methods using visual and range data streams.

Hand gestures are broadly divided into two types: static gestures and dynamic gestures. Static gestures are gestures which involve only a specific hand pose, and can be recognized using only spatial sensor data. Dynamic gestures involve specific movements of the hand, and can be recognized only by considering both spatial and temporal data. Most popular datasets for hand gesture recognition \cite{LDCon_GR, jester_dataset, nv_gesture, egogesture} consist of both static and dynamic gestures.

The state-of-the-art techniques for gesture recognition use deep convolutional neural networks (CNNs) that take a stream of visual and/or range data as input, and finally predict a gesture type.  They are broadly divided into networks which estimate hand poses before predicting the gesture type, and networks which directly classify the input signal into a gesture type. Techniques which estimate a 2D hand pose use 2D CNNs, either to regress the exact co-ordinates of hand keypoints \cite{coordinate_hand_pose1, coordinate_hand_pose2}, or to generate a rough heatmap for hand keypoint locations \cite{heatmap_hand_pose1, heatmap_hand_pose2}. These techniques are less suited to long-distance applications where estimating exact hand keypoints is difficult.

%The pose is used to classify the signal using various techniques like Graph convolution networks\cite{li2019spatial}.
Techniques which directly classify the input make use of 3D CNN architectures \cite{3d_cnn_orig, slowfast1, c3d}(for example 3D ResNets \cite{3D_resnet}) to effectively utilize and fuse both spatial and temporal features for gesture recognition. Some of these techniques have additional modules made of custom CNNs to fuse different modalities\cite{MMTM_gest_Rec}. Other methods \cite{RNN_gest_rec} make use of recurrent neural networks (LSTMs) to model the temporal dependencies, while using CNNs to extract spatial features. Methods based on extracting and utilizing optical flow \cite{flow_gest_Rec} have also been used for hand gesture recognition. 

Some recognition methods \cite{real_time_3d} include a separate light-weight detection module to help decide if the main gesture classifier is to be used on an input data stream or not. Such a light-weight detector improves the efficiency of most of the methods described previously.

Our results suggest that these techniques fail to accurately recognize gestures when the subject is far away ($>1 $ meter) from the capturing system. The gesture occupying only a small spatial region of the sensor’s field-of-view (FOV) is one of the main reasons for this reduction in accuracy.
\begin{figure*}[h!]
	 \centering
    \includegraphics[width=\textwidth]{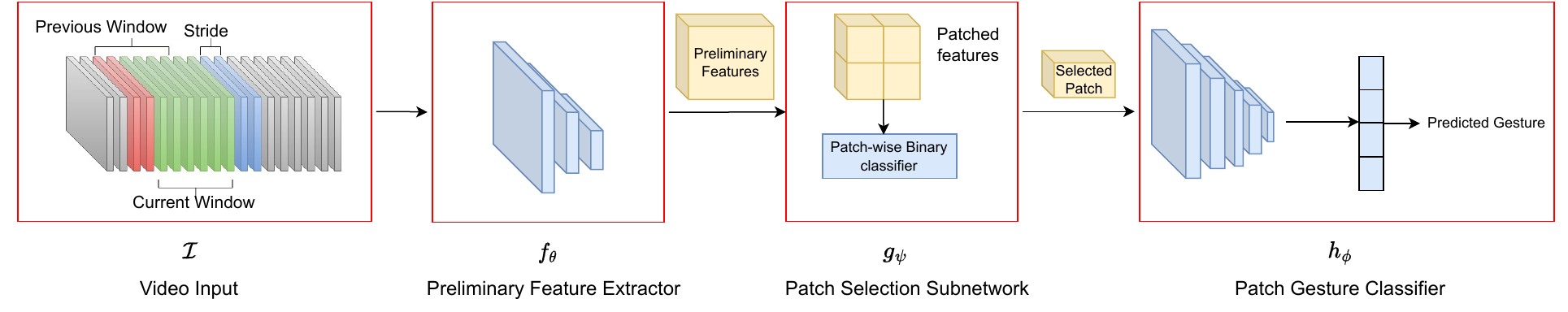}
    \caption{The proposed dynamic neural network consists of 3 blocks: 1) The preliminary feature extractor, 2) Patch selection subnetwork and, 3) The patch gesture classifier.}
    \label{fig:network_block}
\end{figure*}
\subsection{Dynamic Neural Networks}
Most deep neural networks perform inference using a static computational graph and static set of parameters, which remain the same for all inputs. Dynamic neural networks on the other hand, adapt their computational structure based on the input. This helps them be more efficient, interpretable and generalizable than their static counterparts \cite{dynamic_orig1, dynamic_orig2}. 
Broadly, dynamic neural networks are categorized into spatially dynamic and temporally dynamic networks \cite{dynamic_survey}.

Temporally dynamic networks \cite{temporal_dynamic1} are used on sequential data such as videos, text and audio. They use a recurrent neural network (RNN) to decide the computation path for an input sample. Typically in video recognition, the input stream is preprocessed through a series of convolutional layers before being fed to an RNN which decides if this video segment requires further processing or not. 

Spatially dynamic neural networks perform dynamic computation by helping the network focus on more meaningful regions of image and video inputs. They are further subdivided into pixel-wise and region-wise dynamic networks, based on their level of spatial adaptability. Pixel-wise dynamic networks include networks with dynamic sparse convolution, which has been used for increasing the efficiency and performance of image classification \cite{pixel_wise_dynamic1,pixel_wise_dynamic2}. Region-wise dynamic networks include either the use of hard attention with RNN's \cite{region_wise_dynamic1} to choose patches of the input, or the use of other methods like class activation maps for determining most informative regions \cite{region_wise_dynamic2}.
Our proposed network is a type of region-wise spatially dynamic neural network.

\section{Method}\label{sec:method}

Our network is a spatially dynamic neural network, which is patch-wise adaptable. It operates on a continuous stream of visual data (RGB video), taking as input a set of T frames at a time and sliding this window of T frames across the stream. The video input block in Figure \ref{fig:network_block} depicts our network working on such a video stream using the sliding window style of operation. 

Specifically, our network operates on a 4 dimensional input of size $T \times  C \times H \times W$, with $T$ representing the number of frames, $C$ representing the number of channels (usually 3), and $H, \ W$ representing the spatial resolution of the image. For each such input $\mathcal{I}$, our network produces a gesture label $y_{\mathcal{I}} \in N$, where $N$ is the number of gesture types in the dataset. Our network consists of 3 main blocks:
\begin{enumerate}
\item Preliminary feature extractor
\item Patch selection subnetwork
\item Patch gesture classifier
\end{enumerate}
Figure \ref{fig:network_block} illustrates the high level structure of our network, including the connections between these blocks. The input $\mathcal{I}$ first goes to the preliminary feature extractor, which is connected to the patch selection subnetwork, which is followed by the patch gesture classifier. We describe the structure and function of each of these blocks in detail in the following subsection.
\subsection{Structural Blocks}
\begin{figure}[t!]
	 \centering
    \includegraphics[width=0.5\textwidth]{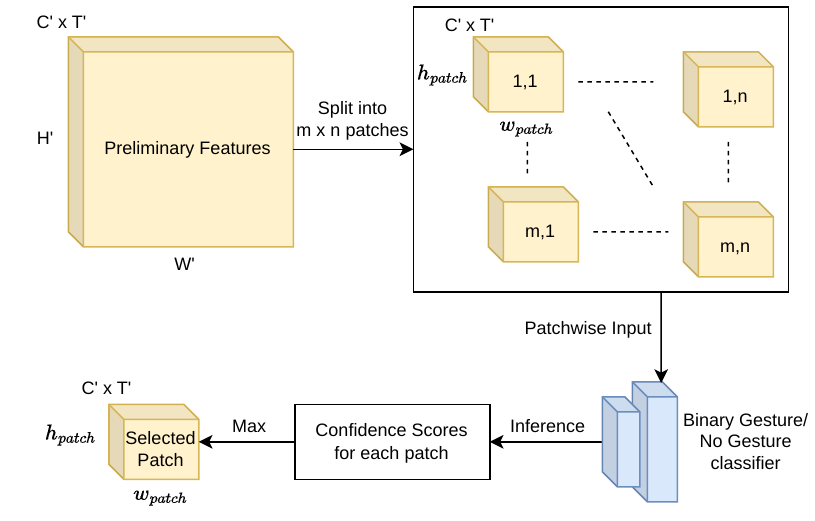}
    \caption{The patch selection subnetwork takes the preliminary features as input, splits them into spatial patches and then selects the patch with the most useful features to be forwarded to the next block of the network.}
    \label{fig:patch_selection}
\end{figure}

\subsubsection{\textbf{Preliminary Feature Extractor}}
This block consists of a shallow network made of 3D convolutional layers acting on the input $\mathcal{I}$. The preliminary feature extractor $f_{\theta}$, parameterized by $\theta$, extracts low level spatio-temporal features $f_{\theta}(\mathcal{I})$ from the full input $\mathcal{I}$. These low level spatio-temporal features are used by the next block to select a patch of features with maximum gesture information. 

The preliminary feature extractor acts on the full spatial input and is hence designed to be shallow to minimize the compute required by it.
\subsubsection{\textbf{Patch Selection Subnetwork}}
The patch selection subnetwork, shown in Figure \ref{fig:patch_selection}, receives low level features $f_{\theta}(\mathcal{I})$ of size $T' \times C' \times H' \times W'$ from the preliminary feature extractor, and divides it into $m \times n$ non-overlapping spatial patches. We denote these patches formed from the full set of features $f_{\theta}(\mathcal{I})$ as $ \{ p_{1,1}, \ldots ,p_{1,n}, p_{2,1}, \ldots ,p_{2,n}, \ldots p_{m,n}  \} $, with each patch having a size $T' \times C' \times h_{patch} \times w_{patch}$. The spatial size of the patches ($h_{patch} \times w_{patch}$) should be large enough to allow a single patch to contain sufficient information to recognize the gesture.

The patch selection subnetwork selects a single patch from these $m \times n$ patches to forward to the next stage for gesture classification. It uses a small 3D CNN $g_{\psi}$, parameterized by $\psi$, as a binary classifier for patch selection. Each patch $p_{i,j}$ is assigned a confidence score $S_{p_{i,j}}$ defined by,
\begin{equation} 
S_{p_{i,j}}=g_{\psi}(p_{i,j}),
\label{eq:patch_conf_score}
\end{equation}
where $g_{\psi}(p_{i,j})$ represents the confidence with which the subnetwork $g_{\psi}$ predicts that the patch $p_{i,j}$ contains a gesture-performing hand. A single feature patch is selected as,
\begin{equation} 
p_{max} \leftarrow \underset{p_{i,j}}{arg\,max}  \ S_{p_{i,j}},
\label{eq:patch_max_assign}
\end{equation}

where $p_{max}$ is the patch associated with the maximum confidence score $S_{max}$. The patch selection greatly reduces the size of the input features forwarded to the most compute intensive block - the patch gesture classifier.

\subsubsection{\textbf{Patch Gesture Classifier}}
The patch gesture classifier $h_{\phi}$, parameterized by $\phi$, consists of 3D convolutional layers which receive features (computed by the preliminary feature extractor) from a single patch selected by the patch selection subnetwork.  The selected input patch $p_{max}$ has a size $T' \times C' \times h_{patch} \times w_{patch}$, for which the gesture classifier outputs a distribution $h_{\phi}(p_{max})$ over $N$ possible gesture classes. The class with the highest probability is predicted as the recognized gesture. The smaller size of the patch features as compared to complete input features (by a factor $\frac{H' \times W'}{h_{patch} \times w_{patch}}$) helps save a significant amount of compute in this part of the network.

The patch gesture classifier can be chosen to be any common 3D convolutional backbone such as 3D ResNet, 3D ResNext, 3D MobileNet \cite{3D_resnet, 3D_mobilenet}, etc, after appropriate modification to receive $C'$ channels of the patched input.

\subsection{Training} 
The whole network, including the patch selection subnetwork and the patch gesture classifier, is trained end-to-end as a single unit. This allows our loss to be expressed as a sum of cross entropy terms, which can be effectively optimized. The loss $\mathcal{L}$ is a sum of the cross entropy loss for gesture recognition, and the cross entropy loss for patch selection. It is defined as,
\begin{equation} 
\mathcal{L}= H(h_{\phi}(p_{max}), y_{\mathcal{I}} ) + \lambda \sum_{i,j} H( g_{\psi}(p_{i,j}), y_{p_{i,j}}).
\label{eq:training_loss}
\end{equation}
Here, $\lambda$ is a relative weight hyperparameter, while $H(a,b)$ depicts the cross entropy between $a$ and $b$. $y_{p_{i,j}}$ is the label for patch $p_{i,j}$ indicating if it has a hand gesture or not.

\section{Experiments and Results}\label{sec:exp}
We evaluate the performance of our method for long-distance gesture recognition using the LD-ConGR dataset \cite{LDCon_GR}. The dataset consists of 542 RGB-D videos at 30 fps with a video resolution of $1280 \times 720$. We use only the RGB channels for our experiments, as inexpensive cameras are more common in robots and consumer devices. Each video has a subject seated in a conference room performing gestures belonging to one or more of 10 gesture categories. The subject is seated at a long-distance ($> 1$ meter and  $< 4$ meters) from the capture device. The gesture categories include both static and dynamic gestures. In addition to frame-wise gesture type annotations in the videos, the dataset also provides rough bounding box annotations of the gesturing hand, which we use to train the patch selection subnetwork. Performance on the dataset is measured using the Top-1 Accuracy for framewise predictions made by the network. 

We use the first 2 layers of the 3D ResNeXt-101 CNN \cite{3D_resnet} as the preliminary feature extractor. The patch selection subnetwork consists of the following sequence of layers with $3 \times 3 \times 3$ 3D convolutions kernels : $conv3d\_bn\_relu(str\ 1)$ $|$ $avg pool(str\ 2)$ $|$ $conv3d\_bn\_relu(str\ 1)$ $|$ $linear$. Our patch gesture classifier is the 3D ResNeXt-101 CNN  \cite{3D_resnet} with $C'$ input channels. The patch gesture classifer is initialized from models trained on the Jester dataset\cite{jester_dataset}. The low level features of size $T'$=16, $C'$=64, $H'$=56, $W'$=56 received by the patch selection subnetwork are divided into $6$ patches. Additionally, following previous work \cite{MMTM_gest_Rec, real_time_3d}, we set $T$=32, $C$=3, $H$=112, $W$=112. The network parameters are trained to optimize the combined loss defined in Equation \ref{eq:training_loss} using SGD with momentum for 70 epochs on 4 NVIDIA V100 GPUs. Initial learning rate was set to $10^{-3}$ and it was decayed using cosine annealing throughout training. The relative weight parameter $\lambda=2$. The above hyperparameters apply to all experiments unless specified otherwise.

\begin{figure*}[h!]
	 \centering
    \includegraphics[width=0.86\textwidth]{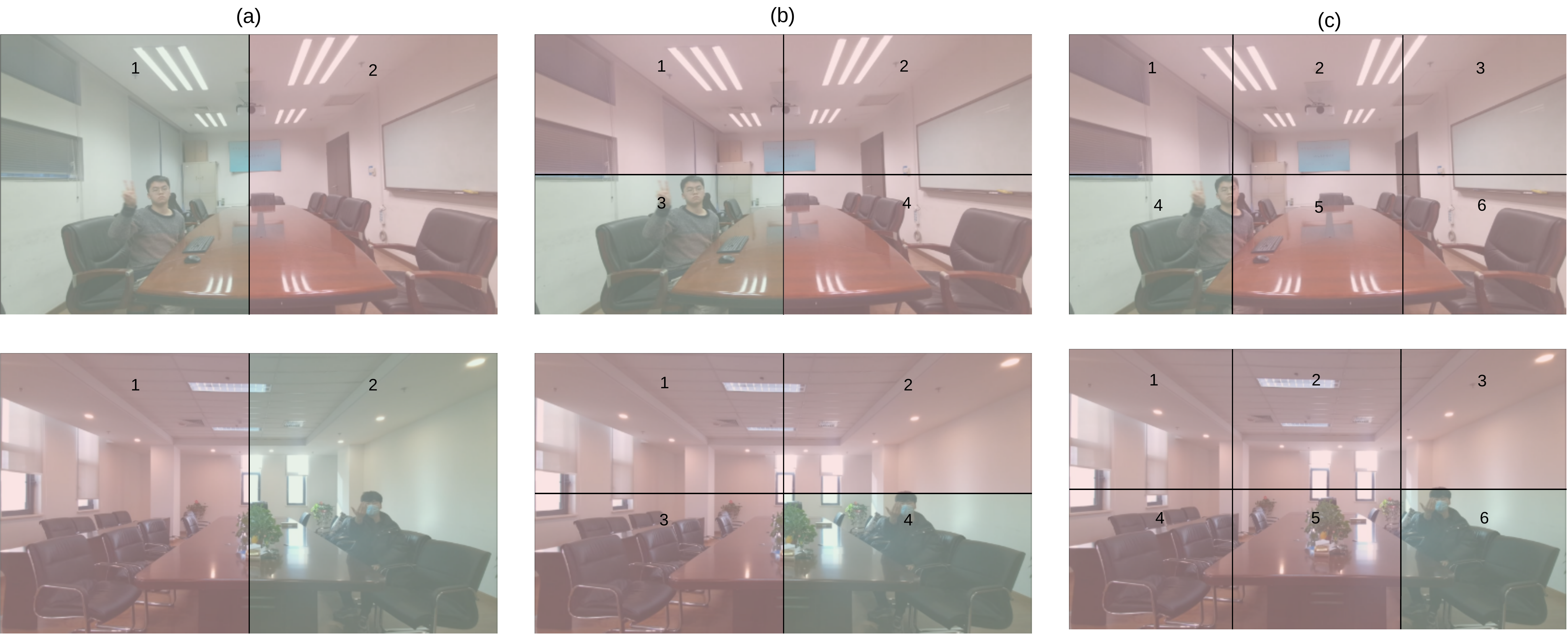}
    \caption{The patch selection subnetwork selects features extracted from the spatial region highlighted in green in the image, and forwards them to the patch gesture classifier. It discards features from parts of the image highlighted in red. The size of the green region can be varied by changing the number of feature patches, as seen in (a) 2 patches,  (b) 4 patches and (c) 6 patches.}
    \label{fig:patch_image}
\end{figure*}
\subsection{Performance on LD-ConGR}
We compare our method with state-of-the-art gesture recognition methods listed below. Due to the absence of hand keypoint annotations in the LD-ConGR dataset, we do not compare with techniques that predict hand pose or keypoints in an intermediate step.
\begin{enumerate}
 \item \textbf{3D ResNeXt-101}\cite{3D_resnet} is the 3D ResNeXt-101 CNN without our dynamic patch selection subnetwork.
 \item \textbf{SlowFast} \cite{slowfast1} uses a combination of high and low frame rate 3D CNNs (based on the 3D ResNeXt-101) for better and more efficient gesture recognition.
 \item \textbf{C3D} \cite{c3d} is a popular 3D CNN baseline for gesture and activity recognition.
 \item \textbf{Temporal Segment Networks} \cite{flow_gest_Rec} use separate networks to extract spatial and temporal features. 
 \end{enumerate}

As can be seen in Table \ref{tab:results_main}, our method achieves state-of-the-art gesture recognition accuracy on LD-ConGR as compared to other methods. It does so while using a significantly lower amount of compute (GFLOPS) than other methods. This is due to the fact that our network discards early on, a bulk of features which have no gesture information.

\begin{table}[h!]
\begin{center}
\begin{tabular}{|c|| c c|}
\hline
\textbf{Method }& \begin{tabular}{@{}c@{}}\textbf{Compute} \\ \textbf{(GFLOPS)} \end{tabular} & \textbf{Top-1  Accuracy ($ \% $)} \\
\hline \hline
Ours & \textbf{10} & \textbf{89.94}\\ \hline
3D ResNeXt-101\cite{3D_resnet} & 16 & 85.33 \\ \hline 
C3D \cite{c3d} & 12 & 82.38 \\ \hline
SlowFast \cite{slowfast1} & 18 & 83.87\\ \hline
TSN  \cite{flow_gest_Rec}& 14 & 86.91 \\ \hline
\end{tabular}
\end{center}
\caption{Our method outperforms other representative baselines on the task of long-distance gesture recognition evaluated on the LD-ConGR dataset}
\label{tab:results_main}
\end{table}
 
\subsection{Performance in a resource constrained environment}
We also evaluate the performance of our method in a resource (power, compute, memory) constrained environment which provides a more realistic indicator of its performance when used for real-world applications like controlling mobile robots and smart home devices. We replace the 3D ResNeXt-101 backbone with the 3D MobileNet\cite{3D_mobilenet}. We use a 4 patch ($2 \times 2$) version of our network, with a single layer deep preliminary feature extractor to ensure that the number of parameters in our network is similar to the 3D MobileNet baseline. This helps ensure that both networks being compared have similar compute requirements and are deployable on resource constrained robots and edge-devices.

Our method is able to vastly improve (by 17.4\%) on the performance of the current state-of-the-art light weight 3D MobileNet CNN, as can be seen in the results in Table \ref{tab:perf_low_res}. C3D \cite{c3d}, Slowfast \cite{slowfast1} and Temporal Segment Networks \cite{flow_gest_Rec} are not included in this comparison because they do not have any comparable lightweight versions available. These results demonstrate the importance of early rejection of background features by our dynamic neural network, helping focus the limited compute on features with gesture information.
\begin{table}[h!]
\begin{center}
\begin{tabular}{|c|| c c|}
\hline
\textbf{Method }& \begin{tabular}{@{}c@{}}\textbf{Compute} \\ \textbf{(GFLOPS)} \end{tabular} & \textbf{Top-1  Accuracy ($ \% $)} \\
\hline \hline
Ours (3D MobileNet) & 1.5 & \textbf{76.68}\\ \hline
3D MobileNet\cite{3D_mobilenet} & 1.5 & 65.33 \\ \hline
\end{tabular}
\end{center}
\caption{Under a constrained compute budget, our method is able to optimally use the compute on features of interest leading to much better performance}
\label{tab:perf_low_res}
\end{table}

\subsection{Analyzing performance for different gestures}
Table \ref{tab:table_gesture_results} tabulates the accuracy achieved by our network and the comparable 3D ResNeXt-101 CNN at recognizing each of the 10 different gestures in the LD-ConGR dataset. Our method outperforms the baseline on most of the gesture types. We observe significant gains in performance for gestures such as Click and Pinch that are relatively difficult to identify, and need finer detection of finger movements. We hypothesize that the gains in performance are due to the early discarding of background features by our method, which helps the patch gesture classifier learn finer features.%, as compared to the baseline 3D ResNeXt-101 CNN.
\begin{table}[h!]
\begin{center}
\begin{tabular}{|c|| c c|}
\hline
\textbf{Classes} & \begin{tabular}{@{}c@{}} \textbf{Accuracy ($ \% $)} \\ \textbf{(Ours)} \end{tabular}  &   \begin{tabular}{@{}c@{}} \textbf{Accuracy ($ \% $)} \\ \textbf{(3D ResNeXt-101)} \end{tabular}  \\
\hline\hline
Pinch & \textbf{63.21} & 38.89 \\ \hline
Click & \textbf{93.45} & 85.45 \\ \hline
Palm & \textbf{85.1} & 84.9 \\ \hline
Fist & \textbf{83.48 }& 76.35 \\ \hline
Thumb up & 95.51 & \textbf{98.08} \\ \hline
Shift Right & \textbf{99.26} & 98.41 \\ \hline
Downward & \textbf{99.10} & 98.89 \\ \hline
Upward & \textbf{94.54} & 93.97 \\ \hline
Left & \textbf{91.46} & 91.28 \\ \hline
Right & 90.86 & \textbf{91.90} \\ \hline
\end{tabular}
\end{center}
\caption{Classification accuracies of our method and the 3D ResNeXt-101 CNN for different gestures in the LD-ConGR dataset}
\label{tab:table_gesture_results}
\end{table}

\begin{table}[h!]
\begin{center}
\begin{tabular}{|c|| c c|}
\hline 
\begin{tabular}{@{}c@{}}\textbf{Number of patches}\\ \textbf{($m \times n$)} \end{tabular}  & \begin{tabular}{@{}c@{}}\textbf{Compute} \\ \textbf{(GFLOPS) }\end{tabular} & \textbf{Top-1 Accuracy ($\% $)}\\
\hline \hline
$1 \times 2$ & 26 & 86.48\\ \hline
$2 \times 2$ & 18 & 88.67 \\ \hline
$2 \times 3$ & 10 & 89.94 \\ \hline
\end{tabular}
\end{center}
\caption{Classification accuracies of our method with different number of patches used in the patch selection subnetwork, showing that a larger number of patches helps improve the network performance}
\label{tab:num_patch}
\end{table}

\subsection{Effect of number of patches}
The number of patches in the incoming preliminary features is an important parameter to be selected. We conduct an experiment to determine its effect on the accuracy and compute efficiency of the network. Specifically, we train separate networks with the features divided into 2, 4 and 6 patches and measure the gesture recognition accuracy achieved by them. We use the ResNeXt-101 backbone in all of these networks. 

The results in Table \ref{tab:num_patch} demonstrate that dividing the input progressively into smaller patches improves the compute efficiency of the network by helping discard more background features early on, while also positively affecting its accuracy. Figure \ref{fig:patch_image} illustrates the regions in the input image from which the selected feature patches were extracted. The region highlighted in green shrinks as the number of patches is increased, hence decreasing the number of features processed by the patch gesture classifier. 

We also observed that though using smaller patches leads to better performance, this trend only holds as long as the chosen feature patches are large enough to fit sufficient gesture information. The preliminary features can also be divided into overlapping patches to increase the recognition accuracy at the cost of increased compute. The target application influences the accuracy VS efficiency tradeoff. %should be made based on the target application.

\subsection{Performance on longer distance videos}
The LD-ConGR test set\cite{LDCon_GR} consists of videos in which the subject is at varying distances (1-4 meters) from the capture device. We conduct an experiment to evaluate the effect of distance on the performance of our method. Specifically, we exclusively test our method on videos from the LD-ConGR test set where the subject is sitting at the far end of the conference room. This corresponds to all videos with the recording spot labels l3 and r3 as defined in \cite{LDCon_GR}. These videos have the subject at the longest distance from the capture device ($\approx 4$ meters). For comparison, we also report the accuracy of other methods on the same set of videos in Table \ref{tab:long_distance}. We observe that the performance of our method shows a much smaller relative deterioration for longer distance videos, than that of other methods.

\begin{table}[h!]
\begin{center}
\begin{tabular}{|l||c  c  c|}
\hline 
\multicolumn{1}{| l ||}{ \textbf{Method}} &\begin{tabular}{@{}c@{}} \textbf{Complete} \\ \textbf{test set} \end{tabular} &\begin{tabular}{@{}c@{}} $\approx$\textbf{ 4 meters} \\ \textbf{subset} \end{tabular} &\begin{tabular}{@{}c@{}} \textbf{Relative} \\ \textbf{deterioration} \end{tabular} \\
\hline \hline
Ours & \textbf{89.94} & \textbf{85.65} & \textbf{4.77} \\ \hline
 3D ResNeXt-101 \cite{3D_resnet}  & 85.33 & 74.78 & 12.36 \\ \hline
C3D \cite{c3d}  & 82.38 & 70.25 & 14.72\\ \hline
SlowFast \cite{slowfast1} & 83.87 & 73.53 & 12.32 \\ \hline
TSN  \cite{flow_gest_Rec} & 86.91 & 81.14 & 7.11 \\ \hline
\end{tabular}
\end{center}
\caption{The top-1 accuracy of our method shows a lower deterioration (\%) for longer distance videos.  }
\label{tab:long_distance}
\end{table}

\section{Conclusion}\label{sec:conclusion}
We present a novel dynamic neural network for the task of long-distance gesture recognition. The early discarding of background features not only helps the network focus on important features required for accurate gesture prediction, but it also reduces the compute requirements and makes the network conducive to deployment in resource constrained settings on robots and edge-devices. We perform extensive experiments and evaluation, and compare against other gesture recognition methods to show the effectiveness of our approach. Our method achieves state-of-the-art performance on the LD-ConGR \cite{LDCon_GR} long-distance gesture dataset. As part of our future work, we intend to explore new neural network architectures that support adaptive patch sizes while being efficient enough to be deployed on resource constrained devices. Another research direction would be the extention of our method for general activity recognition at arbitrary distances while taking in multi-modal sensor inputs. 

\section{Acknowledgement}
This work was partially supported by ONR grant N00014-20-1-2444 and USDA NIFA grant 2020-67021-32799/1024178. We thank Marcus Gualtieri and the anonymous reviewers for providing helpful comments.

%%%%%%%%%%%%%%%%%%%%%%%%%%%%%%%%%%%%%%%%%%%%%%%%%%%%%%%%%%%%%%%%%%%%%%%%%%%%%%%%

\bibliographystyle{IEEEtran}
\bibliography{egbib}
\end{document}